\documentclass[11pt]{article}
\usepackage{acl2016}
\usepackage{times}
\usepackage{latexsym}
\usepackage{url}

\aclfinalcopy 
\def\aclpaperid{847} 

\title{Achieving Open Vocabulary Neural Machine Translation \\ with Hybrid Word-Character Models}

\author{Minh-Thang Luong \and Christopher D. Manning\\
        Computer Science Department, Stanford University, Stanford, CA 94305 \\
	    {\tt \{lmthang,manning\}@stanford.edu}
}

\date{}

\usepackage{caption} 
\usepackage{subcaption} 
\usepackage{multirow} 
\usepackage{amsmath} 
\usepackage{amssymb} 
\usepackage{graphicx}
\usepackage{color} 
\usepackage{pstricks} 
\usepackage{arydshln} 
\usepackage[T1]{fontenc}  

\newcommand{\imgExt}{eps}

\newcommand{\bi}[1]{\textbf{\textit{#1}}}
\newcommand{\source}[1]{\bi{#1}}
\newcommand{\correct}[1]{\bi{\color{blue} #1}}
\newcommand{\wrong}[1]{\textbf{\color{red} #1}}
\newcommand{\close}[1]{\underline{\color{brown} #1}}
\newcommand{\hide}[1]{}
\newcommand{\todo}[1]{{\color{brown} TODO: #1}}
\newcommand{\word}[1]{``#1''}

\newcommand{\tgt}[1]{y_{#1}} 
\newcommand{\src}[1]{x_{#1}} 

\newcommand{\MB}[1]{\mbox{\boldmath{$#1$}}} 
\newcommand{\open}[1]{\left(#1\right)} 
\newcommand{\hd}[1]{\MB{h}_{#1}} 
\newcommand{\hb}[1]{\MB{\bar{h}}_{#1}} 
\newcommand{\eq}[1]{Eq.~(\ref{#1})}
\DeclareMathOperator{\softmax}{softmax}

\newcommand{\hi}{\MB{h}_{t}} 
\newcommand{\hs}{\MB{\tilde{h}}_{t}} 
\newcommand{\co}{\MB{c}_{t}} 
\newcommand{\W}[1]{\MB{W_{#1}}} 

\newcommand{\Wc}{\MB{\breve{W}}} 
\newcommand{\hc}{\MB{\breve{h}}_{t}} 

\newcommand{\unk}{$<${\it unk}$>$}

\newcommand{\modelword}{{\it (d)}}
\newcommand{\modelchar}{{\it (g)}}
\newcommand{\modelsmall}{{\it (k)}}
\newcommand{\model}{{\it (l)}}
\newcommand{\ensbleu}{20.7}
\newcommand{\gain}{2.1{-}11.4}
\newcommand{\chr}{chrF$_3$}

\begin{document}

\maketitle

\begin{abstract}
Nearly all previous work on neural machine translation (NMT) has used quite restricted
vocabularies, perhaps with a subsequent method to patch in unknown words. This
paper presents a novel word-character solution to achieving open vocabulary NMT. 
We build hybrid systems that translate mostly at the {\it word}
level and consult the {\it character} components for rare words. 
Our character-level recurrent neural networks compute source
word representations and recover unknown target words when needed.
The twofold advantage of such a hybrid approach is that it is much faster and easier to
train than character-based ones; at the same time, it never produces unknown words as in the case of word-based models. 
On the WMT'15 English to Czech translation task, 
this hybrid approach offers an addition boost of +$\gain{}$ BLEU points over models 
that already handle unknown words. 
Our best system achieves a new state-of-the-art result with
$\ensbleu{}$ BLEU score.
We demonstrate that our character models can successfully learn to not only generate well-formed words for Czech, a
highly-inflected language with a very complex vocabulary, but also build correct
representations for English source words.
\end{abstract}

\section{Introduction}
\label{sec:intro}
Neural Machine Translation (NMT) is a simple new architecture for getting
machines to translate. At its core, NMT is a single deep 
neural network that is trained end-to-end with several advantages such as
simplicity and generalization. Despite being relatively new, NMT has already
achieved state-of-the-art translation results for several language pairs 
such as English-French \cite{luong15}, English-German
\cite{jean15,luong15attn,luong15iwslt}, and English-Czech \cite{jean15wmt}. 
\begin{figure}
\centering
\includegraphics[width=0.4\textwidth, clip=true, trim= 0 0 0 0]{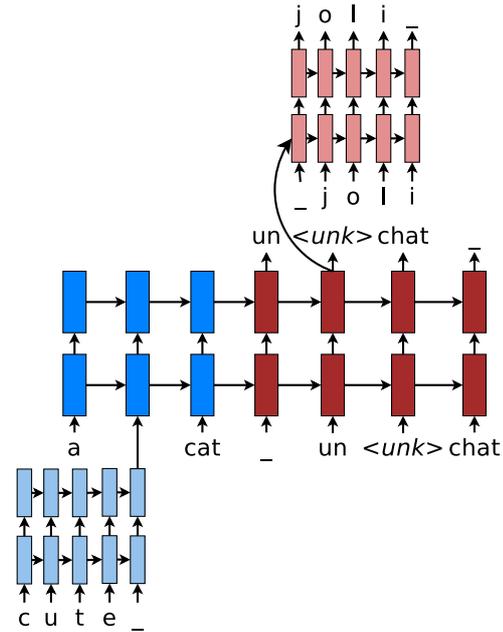}
\caption{{\bf Hybrid NMT} -- example of a word-character model for translating
\word{a cute cat} into \word{un
joli chat}. Hybrid NMT translates at the word level. For rare tokens,
the character-level components build source representations
and recover target \unk{}. \word{\_} marks sequence
boundaries.}
\label{f:hybrid}
\end{figure}

While NMT offers many advantages over traditional phrase-based approaches, such as
small memory footprint and simple decoder implementation, nearly all previous
work in NMT has used quite restricted vocabularies, crudely treating all other
words the same with an \unk{} symbol. Sometimes, a post-processing step that
patches in unknown words is introduced to alleviate this problem. 
\newcite{luong15} propose to annotate occurrences of target \unk{} with positional information to
track their alignments, after which simple word dictionary
lookup or identity copy can be performed to replace \unk{} in the translation.
\newcite{jean15} approach the problem similarly but obtain the alignments for unknown
words from the attention mechanism. We refer to these as the {\it
unk replacement} technique.

Though simple, these approaches ignore several important
properties of languages. First, {\it monolingually}, words are morphologically
related; however, they are currently treated as independent entities. This is
problematic as pointed out by
\newcite{luong13}: neural networks can learn good
representations for frequent words such as \word{distinct}, but fail for
rare-but-related words like \word{distinctiveness}. Second, {\it crosslingually},
languages have different alphabets, so one cannot na\"{i}vely memorize all
possible surface word translations such as name transliteration between 
\word{Christopher} (English) and \word{Kry\u{s}tof} (Czech). See more on this problem
in \cite{sennrich16sub}.

To overcome these shortcomings, we propose a novel {\it hybrid} architecture for NMT
that translates mostly at the word level and consults the character
components for rare words when necessary. As illustrated in
Figure~\ref{f:hybrid}, our hybrid model consists of a word-based NMT that
performs most of the translation job, except for the two (hypothetically) rare words,
\word{cute} and \word{joli}, that are handled separately. On the {\it source}
side, representations for rare words, \word{cute}, are
computed on-the-fly using a deep recurrent neural network that operates at the
character level. On the {\it target} side, we have a separate model that
recovers the surface forms, \word{joli}, of \unk{} tokens character-by-character.
These components are learned jointly end-to-end, removing the need for a separate
unk replacement step as in current NMT practice.

Our hybrid NMT offers a twofold advantage: it is much faster and easier to
train than character-based models; at the same time, it never produces unknown
words as in the case of word-based ones.
We demonstrate at scale that on the WMT'15 English to
Czech translation task, such a hybrid approach provides
an additional boost of +$\gain{}$ BLEU points over models 
that already handle unknown words.
We achieve a new state-of-the-art result with
$\ensbleu{}$ BLEU score.
Our analysis demonstrates that our character models can successfully learn to not
only generate well-formed words for Czech, a
highly-inflected language with a very complex vocabulary, but also build correct
representations for English source words.

We provide code, data, and models at \url{http://nlp.stanford.edu/projects/nmt}.

\section{Related Work}
There has been a recent line of work on end-to-end character-based neural models
which achieve good results for part-of-speech tagging \cite{santos14,ling15function},
dependency parsing \cite{ballesteros15}, text classification
\cite{zhang15}, speech recognition \cite{chan16,bahdanau16}, and language
modeling \cite{kim16,rafal16}. However, success has not been shown for
cross-lingual tasks such as machine translation.\footnote{Recently,
\newcite{ling15char} attempt character-level NMT; however,
the experimental evidence is weak. The authors demonstrate only small
improvements over word-level baselines and acknowledge that there are no differences of
significance. Furthermore, only small datasets were used without
comparable results from past NMT work.}
\newcite{sennrich16sub} propose to segment words into smaller units and
translate just like at the word level, which does not learn to understand
relationships among words.

Our work takes inspiration from \cite{luong13} and 
\cite{li15}. Similar to the former, we build representations for rare words
on-the-fly from subword units. However, we utilize recurrent neural networks
with characters as the basic units; whereas \newcite{luong13} use recursive neural
networks with morphemes as units, which requires existence of a
morphological analyzer. In comparison with \cite{li15}, our hybrid architecture
is also a hierarchical sequence-to-sequence model, but operates at a different
granularity level, word-character. In contrast, \newcite{li15} build
hierarchical models at the sentence-word level for paragraphs and documents.

\section{Background \& Our Models}
\label{sec:nmt}
Neural machine translation aims to directly model the conditional probability $p(\tgt{}|\src{})$ of translating
a source sentence, $\src{1},\ldots,\src{n}$, to a target sentence, $\tgt{1},\ldots,\tgt{m}$.
It accomplishes this goal through an {\it encoder-decoder} framework
\cite{kal13,sutskever14,cho14}. The {\it encoder} computes a representation $\MB{s}$
for each source sentence. Based on that source representation,
the {\it decoder} generates a translation, one target word at a time, and hence,
decomposes the log conditional probability as:
\begin{equation}
\log p(\tgt{}|\src{}) = \sum_{t=1}^m \nolimits \log
p\open{\tgt{t}|\tgt{<t},\MB{s}}
\label{e:s2s}
\end{equation}

A natural model for sequential data is the recurrent
neural network (RNN), used by most of the recent NMT work.
Papers, however, differ in terms of: (a) architecture -- from unidirectional, to
bidirectional, and deep multi-layer RNNs; and (b) RNN type -- which are long
short-term memory (LSTM)
\cite{lstm97} and the gated recurrent unit \cite{cho14}. 
All our models utilize the {\it deep multi-layer} architecture with {\it
LSTM} as the recurrent unit; detailed formulations are in \cite{zaremba14}.

\begin{figure}
\centering
\rput(4.3,6.5){$\tgt{t}$}
\rput(1.4,5.6){$\co$}
\rput(0.8,3.1){$\hb{1}$}
\rput(2.9,3.1){$\hb{n}$}
\rput(4.9,3.1){$\hd{t}$}
\rput(4.9,6.0){$\hs$}
\includegraphics[width=0.33\textwidth, clip=true, trim= 0 0 0 0]{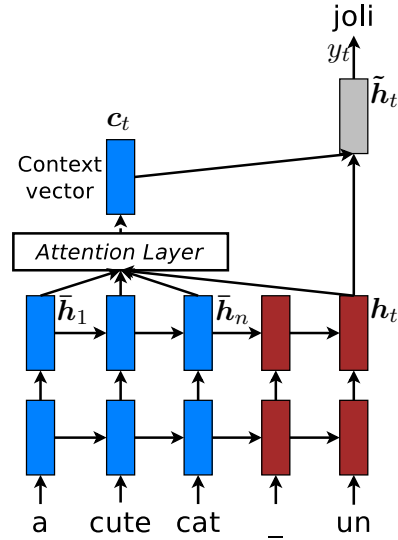} 
\caption{{\bf Attention mechanism}.
} 
\label{f:attn}
\end{figure}

Considering the top recurrent layer in a deep LSTM, 
with $\hd{t}$ being the current target hidden state as in Figure~\ref{f:attn}, one can compute the probability of decoding each target word $y_t$ as:
\begin{equation}
p\left(\tgt{t}|\tgt{<t},\MB{s}\right) = \softmax\open{\hd{t}}
\label{e:softmax}
\end{equation}

For a parallel corpus $\mathbb{D}$, we train our model by minimizing the below
cross-entropy loss:
\begin{equation}
J = \sum_{(\src{},\tgt{}) \in \mathbb{D}} \nolimits -\log p(\tgt{}|\src{})
\label{e:obj}
\end{equation}

\noindent {\bf Attention Mechanism} -- 
The early NMT approaches \cite{sutskever14,cho14}, which we have described above, use only the last encoder state
 to initialize the decoder, i.e., setting the input representation $\MB{s}$ in
 \eq{e:s2s} to $[\hb{n}]$. Recently, \newcite{bog15} propose an {\it attention
 mechanism}, a form of random access memory for NMT to 
cope with long input sequences.
\newcite{luong15attn} further extend the attention mechanism to different
scoring functions, used to compare source and target
hidden states, as well as different
strategies to place the attention.
In all our models, we utilize the {\it global} attention mechanism and the {\it
bilinear form} for the attention scoring function similar to 
\cite{luong15attn}.

Specifically, we set $\MB{s}$ in \eq{e:s2s} to
the set of source hidden states at the top layer, $[\hb{1}, \dots, \hb{n}]$. 
As illustrated in Figure~\ref{f:attn}, the attention mechanism consists of two stages: (a) {\it
context vector} -- the current hidden state $\hd{t}$ is compared with
individual source hidden states in $\MB{s}$ to learn an alignment vector, which
is then used to compute the context vector $\co$ as a weighted average of
$\MB{s}$; and (b) {\it
attentional hidden state} -- the context vector $\co$ is then used to derive a
new attentional hidden state:
\begin{equation}
\hs = \tanh(\W{}[\co; \hi])
\label{e:hs}
\end{equation} 
The attentional vector $\hs$ then replaces $\hd{t}$ in \eq{e:softmax} in
predicting the next word.

\section{Hybrid Neural Machine Translation}
\label{sec:hybrid}
Our hybrid architecture, illustrated in Figure~\ref{f:hybrid}, leverages the power of both words
and characters to achieve the goal of open vocabulary NMT. The core of the
design is a {\it word}-level NMT with the advantage of being fast and easy to
train.
The {\it character} components empower the 
word-level system with the abilities to compute any source word representation on the fly from 
characters and to recover character-by-character unknown target words
originally produced as \unk{}.

\subsection{Word-based Translation as a Backbone}
\label{subsec:hybrid_word}

The core of our hybrid NMT is a deep LSTM encoder-decoder that translates at
the {\it word} level as described in Section~\ref{sec:nmt}. We maintain a
vocabulary of $|V|$ frequent words for each language. Other words not inside these
lists are represented by a universal symbol \unk{}, one per language.
We translate just like a word-based NMT system with respect to these source and
target vocabularies, except for cases that involve \unk{} in the source input or 
the target output. These correspond to the character-level components 
illustrated in Figure~\ref{f:hybrid}.

A nice property of our hybrid approach is that by varying the vocabulary size,
 one can control how much to blend the
word- and character-based models; hence, taking the best of both
worlds. 

\subsection{Source Character-based Representation}
\label{subsec:src}
In regular word-based
NMT, for all rare words outside the source vocabulary, one feeds the
universal embedding representing \unk{} as input to the encoder. This is
problematic because it discards valuable information about the source word. To
fix that, we learn a deep LSTM model over characters
of source words. 
For example, in Figure~\ref{f:hybrid}, we run
our deep character-based LSTM over `c', `u', `t', `e', and `\_' (the boundary
symbol). The final hidden state at the top layer will be used as the on-the-fly
representation for the current rare word.

The layers of the deep character-based LSTM are always initialized with {\it
zero} states. One might propose to connect hidden
states of the word-based LSTM to the character-based model; however, we chose this design
for various reasons. First, it simplifies the architecture. Second, it allows
for efficiency through {\it precomputation}: before each mini-batch, we can compute
representations for rare source words all at once. All instances of the same
word share the same embedding, so the computation is per {\it type}.\footnote{While \newcite{ling15char} found that it is slow and difficult to train
source character-level models and had to resort to pretraining, we demonstrate
later that we can train our deep character-level LSTM 
perfectly fine in an end-to-end fashion.} 



\subsection{Target Character-level Generation}
\label{subsec:tgt}
General word-based NMT allows generation of \unk{} in the target output.
Afterwards, there is usually a post-processing step that handles
these unknown tokens by utilizing the
alignment information derived from the attention mechanism and then performing
simple word dictionary lookup or identity copy \cite{luong15attn,jean15}. 
While this approach works, it suffers from various problems such as alphabet
mismatches between the source and target vocabularies and multi-word
alignments.
Our goal is to address all
these issues and create a
coherent framework that handles an unlimited output vocabulary.

Our solution is to have a separate deep LSTM that ``translates'' at the
character level given the current word-level state. We train our system such
that whenever the word-level NMT
produces an \unk{}, we can consult this character-level decoder to recover the correct surface form of the
unknown target word. This is illustrated in Figure~\ref{f:hybrid}.
The training objective in \eq{e:obj} now becomes: 
\begin{equation}
J = J_{w} + \alpha J_{c}
\label{e:char_obj}
\end{equation}
Here, $J_{w}$ refers to the usual loss of the word-level NMT; in
our example, it is the sum of the negative log likelihood of
generating $\{\mbox{``un'', ``\unk{}'', ``chat'', ``\_''}\}$. The remaining component $J_c$
corresponds to the loss incurred by the character-level 
decoder when predicting characters, e.g., $\{\mbox{`j', `o', `l', `i',
`\_'}\}$, of those rare words not in the
target vocabulary. 

\paragraph{Hidden-state Initialization} 
\label{subsubsec:h}
Unlike the source character-based representations, which are
context-independent, the target character-level generation requires the
current word-level context to produce meaningful translation.
This brings up an important
question about what can best represent the current context so as to
initialize the character-level decoder. We answer this question in the context
of the attention mechanism ($\S$\ref{sec:nmt}). 

The final vector $\hs$, just before the
softmax as shown in Figure~\ref{f:attn}, seems to be a good candidate to initialize the character-level decoder.
The reason is that $\hs$ combines
information from both the context vector $\co$ and the top-level recurrent
state $\hi$. We refer to it later in our
experiments as the \textit{same-path} target generation approach.

On the other hand, the same-path approach worries us because all vectors $\hs$
used to seed the character-level decoder might have similar values, leading to
the same character sequence being produced.
The reason is because $\hs$ is directly used in the softmax, \eq{e:softmax}, to predict the same \unk{}.
That might pose some challenges for the model to learn useful representations
that can be used to accomplish two tasks at the same time, that is to predict
\unk{} and to generate character sequences.
To address that concern, we propose another approach called
the \textit{separate-path} target generation.

Our separate-path target generation approach works as follows. We mimic the
process described in \eq{e:hs} to create a counterpart vector $\hc$ that will be
used to seed the character-level decoder:
\begin{equation}
\hc = \tanh(\Wc[\co; \hi])
\label{e:hc}
\end{equation} 
Here, $\Wc$ is a new learnable parameter matrix, with which we
hope to release $\W{}$ from the pressure of having to extract information
relevant to both the word- and character-generation processes.
Only the hidden
state of the first layer is initialized as discussed above. The other components
in the character-level decoder such as the
LSTM cells of all layers and the hidden states of higher layers, all start with zero values.


Implementation-wise, the computation in the
character-level decoder is done per word {\it token} instead of per {\it type} as in the source
character component ($\S$\ref{subsec:src}). 
This is because of the context-dependent nature of the decoder.



\paragraph{Word-Character Generation Strategy}
\label{subsubsec:strategy}
With the character-level decoder, we can view the final hidden states as representations for
the surface forms of unknown tokens and could have fed these to the next
time step. However, we chose not to do so for the efficiency reason explained
next; instead, \unk{} is fed to the word-level decoder
``as is'' using its corresponding word embedding.

During {\it training}, this design choice decouples all executions over \unk{} instances of the
character-level decoder as soon the word-level NMT
completes. As such, the forward and backward passes of the character-level
decoder over rare words can be invoked in batch mode. At {\it test} time,
our strategy is to first run a beam search decoder at the word level to
find the best translations given by the
word-level NMT. Such translations contains \unk{} tokens, so we utilize our
character-level decoder with beam search to generate actual words for these \unk{}.

\section{Experiments}
\label{sec:exp}
We evaluate the effectiveness of our models on the publicly available WMT'15
translation task from English into Czech with 
{\it newstest2013} (3000 sentences) as
a development set 
and {\it newstest2015} (2656 sentences) as a test set. Two metrics are used: case-sensitive NIST BLEU \cite{Papineni02bleu}
and \chr{} \cite{chrf}.\footnote{For NIST BLEU, we first run
\texttt{detokenizer.pl} 
and then use \texttt{mteval-v13a}
to compute the scores as per WMT guideline. For \chr{}, we utilize the implementation here
\url{https://github.com/rsennrich/subword-nmt}.}
The latter measures the amounts of overlapping character $n$-grams and has
been argued to be a better metric for translation tasks out of English.

\subsection{Data}
Among the available language pairs in WMT'15, all involving English, 
we choose {\it Czech} as a target language for several reasons. First and
foremost, Czech is a Slavic language with not only rich
and complex inflection,
but also fusional morphology in which a single morpheme can encode multiple
grammatical, syntactic, or semantic meanings. As a result, Czech possesses an enormously large
vocabulary (about 1.5 to 2 times bigger than that of English according to 
statistics in Table~\ref{t:data}) and is a challenging language to translate
into. Furthermore, this language pair has a large
amount of training data, so 
we can evaluate at scale. Lastly, though our techniques are language
independent, it is easier for us to work with Czech since Czech uses the Latin alphabet with some
diacritics. 

\begin{table} 
\centering
\resizebox{8cm}{!}{
\begin{tabular}{l|c|c|c|c}
& \multicolumn{2}{c|}{\bf{English}} & \multicolumn{2}{c}{\bf{Czech}}\\
  \cline{2-5}
& word & char & word & char \\
  \hline
  \# Sents & \multicolumn{4}{c}{15.8M} \\
  \hdashline
  \# Tokens & 254M & 1,269M & 224M & 1,347M \\ 
 \hdashline
  \# Types & 1,172K & 2003 & 1,760K & 2053\\ 
  \hline
  200-char & \multicolumn{2}{c|}{98.1\%} & \multicolumn{2}{c}{98.8\%} \\
\end{tabular}
}
\caption{{\bf WMT'15 English-Czech data} -- shown are various statistics of our training
data such as {\it sentence}, {\it token} (word and character counts), as well as
{\it type} (sizes of the word and character vocabularies).
We show in addition the amount of words in a vocabulary expressed by a list of 200 characters found
in frequent words.}
\label{t:data}
\end{table}

In terms of preprocessing, we apply only the standard tokenization practice.\footnote{Use \texttt{tokenizer.perl} in Moses with
default settings.} We choose for each language a list of 200
characters found in frequent words, which, as shown in Table~\ref{t:data}, can
represent more than 98\% of the vocabulary. 


\begin{table*}
\centering
\begin{tabular}{c|l|c|c|c|c|c}
 & \multirow{2}{*}{\bf{System}} & \multirow{2}{*}{{\bf
 Vocab}} &
\multicolumn{2}{c|}{{\bf Perplexity}} &\multirow{ 2}{*}{\bf{BLEU}} &\multirow{
2}{*}{\bf{\chr{}}}\\
\cline{4-5}
& & & w & c & & \\
  \hline
(a) & Best WMT'15, big data \cite{bojar15wmt} & 
- & - & - & \bi{18.8} & - \\
  \hline
\multicolumn{7}{c}{{\it Existing} NMT}\\
  \hline
(b) & RNNsearch + unk replace \cite{jean15wmt} & 200K & - & - & 15.7 & - \\
(c) & \bi{Ensemble} 4 models + unk replace \cite{jean15wmt} & 200K & - & - & 18.3 & - \\
  \hline
\multicolumn{7}{c}{Our {\it word-based} NMT}\\
  \hline
(d) & Base + attention + unk replace & 50K & 5.9 & - & 17.5 & 42.4 \\
(e) & \bi{Ensemble} 4 models + unk replace & 50K & - & - & 18.4 & 43.9 \\
  \hline
\multicolumn{7}{c}{Our {\it character-based} NMT}\\
  \hline
(f) & Base-512 (600-step backprop) & 200 & - & 2.4 & 3.8 & 25.9\\
(g) & Base-512 + attention (600-step backprop) & 200 & - & 1.6 & 17.5 &
\bi{46.6} \\
  \hdashline
(h) & Base-1024 + attention (300-step backprop) & 200 & - & 1.9 & 15.7 & 41.1 \\
  \hline
\multicolumn{7}{c}{Our {\it hybrid} NMT}\\
  \hline
(i) & Base + attention + same-path & 10K & 4.9 & 1.7 & 14.1 & 37.2 \\
(j) & Base + attention + separate-path & 10K & 4.9 & 1.7 & 15.6 & 39.6 \\
(k) & Base + attention + separate-path + 2-layer char & 10K & 4.7 & 1.6 & \bi{17.7} & 44.1 \\
  \hdashline
(l) & Base + attention + separate-path + 2-layer char & 50K & 5.7 & 1.6 & 19.6 & 46.5 \\
(m) & \bi{Ensemble} 4 models & 50K & - & - & {\bf \ensbleu{}} & {\bf 47.5} \\
\end{tabular}
\caption{{\bf WMT'15 English-Czech results} -- shown are 
the vocabulary sizes, perplexities, BLEU, and \chr{} scores of various systems on
{\it newstest2015}. Perplexities are listed under two
categories, word (w) and character (c). 
{\bf Best} and
\bi{important} results per
metric are highlighed.
}
\label{t:encs}
\end{table*}

\subsection{Training Details}
We train three types of systems, purely {\it word-based}, purely {\it
character-based}, and {\it hybrid}.
Common to these architectures is a word-based NMT since the
character-based systems are essentially word-based ones with
longer sequences and the core of hybrid models is also a word-based NMT.

In training word-based NMT, we follow \newcite{luong15attn} to use the global attention mechanism together with
similar hyperparameters: (a) deep LSTM models, 4 layers, 1024
cells, and 1024-dimensional embeddings, (b) uniform initialization of
parameters in $[-0.1, 0.1]$, (c) 6-epoch training with plain SGD and a simple learning
rate schedule -- start with a learning rate of $1.0$; after 4 epochs,
halve the learning rate every 0.5 epoch, (d) mini-batches are of
size 128 and shuffled, (e) the gradient is rescaled whenever its norm exceeds 5, and (f)
dropout is used with probability $0.2$ according to 
\cite{pham2014dropout}.
We now detail differences across the three architectures.

{\bf Word-based NMT} -- We constrain our source and target sequences to
have a maximum length of 50 each; words that go past the boundary are ignored.
The vocabularies are limited to the top $|V|$ most 
frequent words in both languages. Words not in these vocabularies
are converted into \unk{}. After translating, we will perform
dictionary\footnote{Obtained from the alignment links produced by the Berkeley
aligner \cite{liang06alignment} over
the training corpus.} lookup or
identity copy for \unk{} using the alignment information from the
attention models. Such procedure is referred as the {\it unk replace}
technique \cite{luong15,jean15}.

{\bf Character-based NMT} -- The source and
target sequences at the character level are often about 5 times longer than their counterparts in the
word-based models as we can infer from the statistics in
Table~\ref{t:data}. Due to memory constraint in GPUs, we limit our source and
target sequences to a maximum length of 150 each, i.e., we backpropagate
through at most 300 timesteps from the decoder to the encoder. With
smaller 512-dimensional models, we can afford to have longer sequences with up
to 600-step backpropagation. 

{\bf Hybrid NMT} -- The {\it word}-level component uses the
same settings as the purely word-based NMT. For the {\it character}-level source
and target components, we experiment with both shallow and deep 1024-dimensional models of
1 and 2 LSTM layers. 
We
set the weight $\alpha$ in \eq{e:char_obj} for our character-level loss to
$1.0$.

{\bf Training Time} -- It takes about 3 weeks to train a word-based model with
$|V|=50K$ and about 3 months to train a character-based model. Training and
testing for the hybrid models are about 10-20\% slower than those of the word-based
models with the same vocabulary size.

\subsection{Results}

We compare our models with several strong systems. These include the
winning entry in WMT'15, which was
trained on a much larger amount of data, 52.6M parallel
 and 393.0M monolingual sentences \cite{bojar15wmt}.\footnote{This
entry combines two independent
systems, a phrase-based Moses model and a deep-syntactic transfer-based model.
Additionally, there is  an automatic
post-editing system with hand-crafted rules to correct errors
in morphological agreement and semantic meanings, e.g., loss of negation.}
In contrast, we merely use the
provided parallel corpus of 15.8M sentences. 
For NMT, to the best of our knowledge, \cite{jean15wmt} has
the best published performance on English-Czech.

As shown in Table~\ref{t:encs}, for a purely {\it word-based} approach, 
our single NMT model outperforms the best single model in \cite{jean15wmt} by
+$1.8$ points despite
using a smaller vocabulary of only 50K words versus 200K words. 
Our ensemble system {\it (e)} slightly outperforms the best previous NMT system with $18.4$ BLEU.

To our surprise, purely {\it character-based} models, though extremely slow to
train and test, perform quite well. The $512$-dimensional attention-based model \modelchar{} is
best, surpassing the single word-based model in
\cite{jean15wmt} despite having much fewer parameters. It even outperforms most NMT
systems  
on \chr{} with $46.6$ points. This indicates that this model translate words that closely but
not exactly match the reference ones as evidenced in
Section~\ref{subsec:samples}. 
We notice two interesting observations. First,
attention is critical for character-based models to work as is obvious from the
poor performance of the non-attentional model; this has also been shown in speech
recognition \cite{chan16}. Second, long time-step backpropagation is more important
as reflected by the fact that the larger $1024$-dimensional model {\it (h)} with shorter
backprogration is inferior to \modelchar{}. 


Our {\it hybrid} models achieve the best results. 
At 10K words, we demonstrate that our {\it
separate-path} strategy for the character-level target generation
($\S$\ref{subsubsec:h}) is effective, yielding an improvement of +$1.5$ BLEU
points when comparing systems {\it (j)} vs. {\it (i)}. A {\it deeper} character-level architecture of 2 LSTM
layers provides another significant
boost of +$2.1$ BLEU. 
With $17.7$ BLEU points, our hybrid system \modelsmall{} has
surpassed word-level NMT models.

When extending to 50K words, we further improve the translation quality.
Our best single model, system \model{} with $19.6$ BLEU, is already better than all
existing systems.
Our ensemble model {\it (m)} further advances the SOTA
result to \bi{\ensbleu} BLEU, outperforming
the winning entry in the WMT'15 English-Czech translation task by a large margin
of +$1.9$ points. Our ensemble model is also best in terms of \chr{} with \bi{47.5} points.

\begin{figure}
\centering
\includegraphics[width=0.5\textwidth, clip=true, trim= 0 0 0 0]{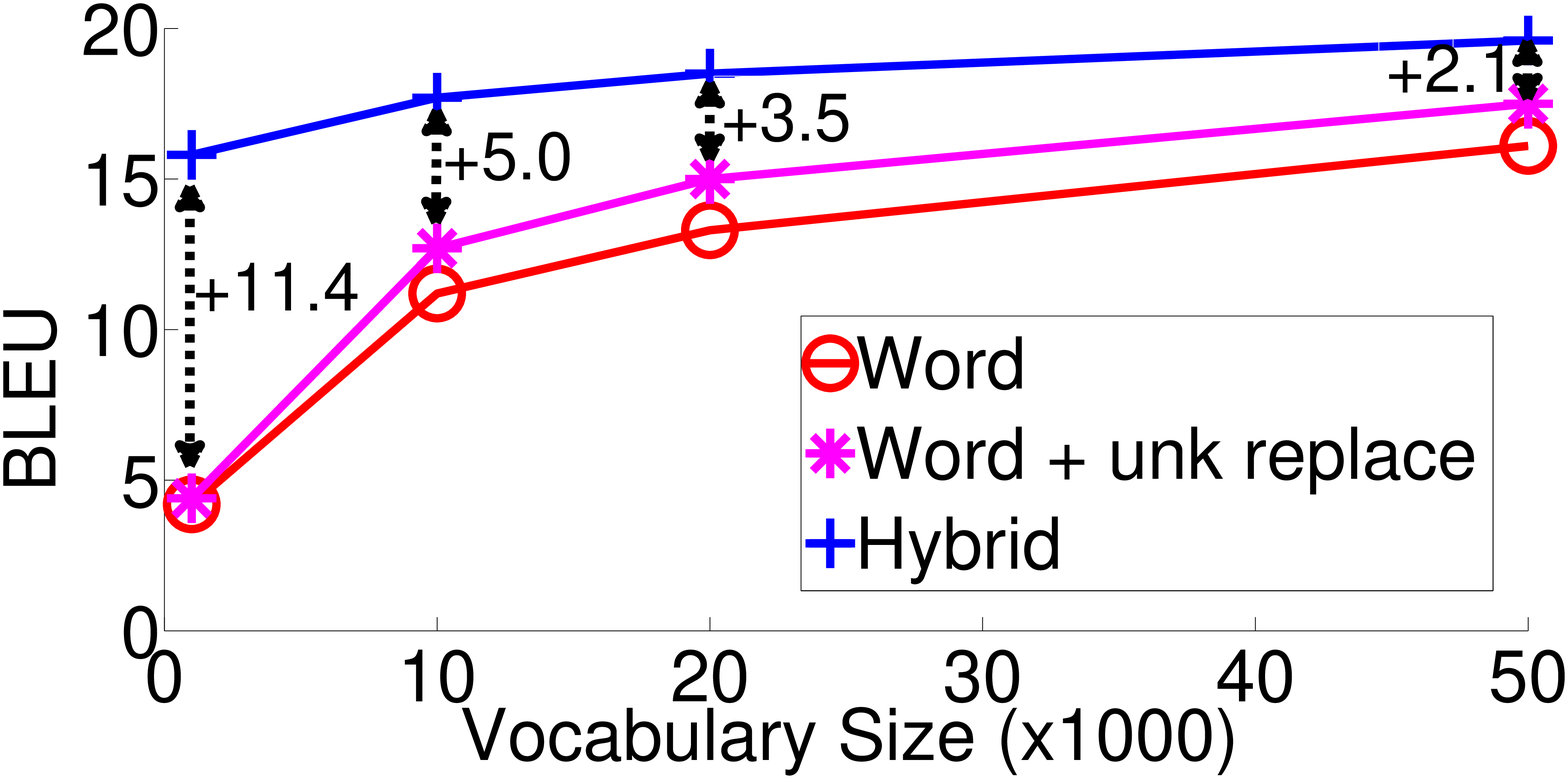}
\caption{{\bf Vocabulary size effect} -- shown are the performances of different
systems as we vary their vocabulary sizes. We highlight the improvements obtained
by our hybrid models over word-based systems which already handle unknown words.}
\label{f:vocab}
\end{figure}

\begin{figure*}
\centering
\includegraphics[width=\textwidth, clip=true, trim= 100 50 0 20]{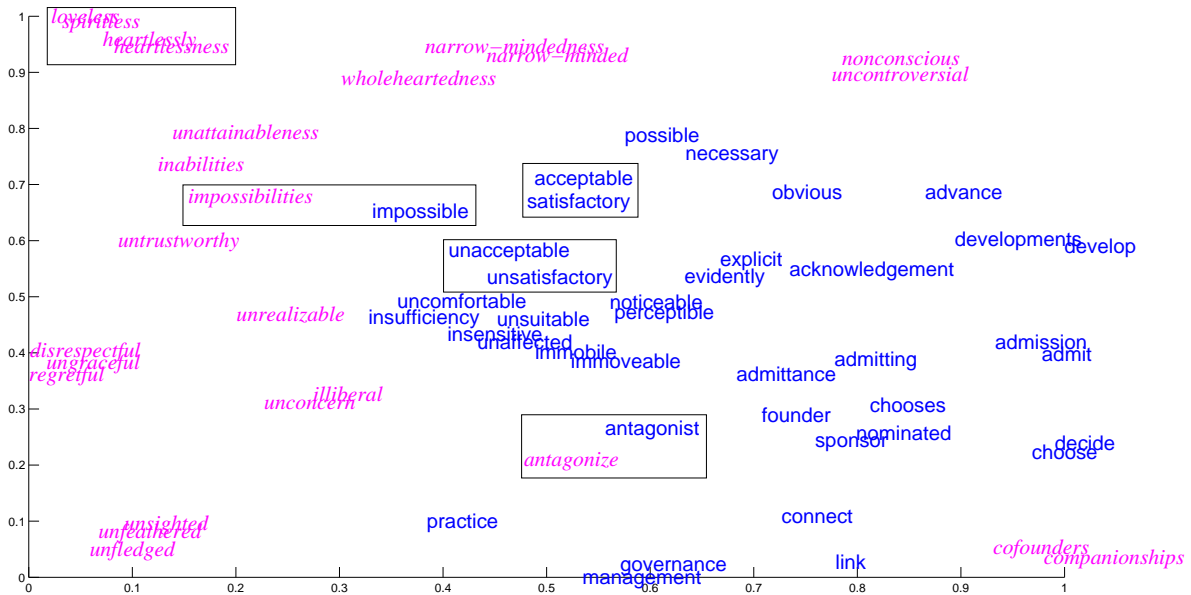}
\caption{{\bf Barnes-Hut-SNE visualization of source word representations} --
shown are sample words from the {\it Rare Word} dataset. We differentiate two types of
embeddings: {\color{blue} frequent} words in which encoder embeddings are looked up directly and {\it {\color{magenta} rare}} words
where we build representations from characters. Boxes highlight examples that
we will discuss in the text. We use the hybrid model \model{} in this visualization.}
\label{f:visual}
\end{figure*}

\section{Analysis}
\label{sec:analysis}
This section first studies the effects of vocabulary sizes towards
translation quality. We then analyze more carefully 
our character-level components by visualizing and evaluating rare word
embeddings as well as examining sample translations.

\subsection{Effects of Vocabulary Sizes}
As shown in Figure~\ref{f:vocab}, our hybrid models offer large gains of
+\gain{} BLEU points over strong word-based systems which already handle unknown words.
With only a small vocabulary, e.g., 1000 words, our hybrid approach can produce
systems that are better than word-based models that possess much larger
vocabularies. While it appears from the plot that gains diminish as we
increase the vocabulary size, we argue that our hybrid models are still
preferable since they understand word structures and can handle new complex
words at test time as illustrated in Section~\ref{subsec:samples}.

\subsection{Rare Word Embeddings}
We evaluate the {\it source} character-level model by building representations
for rare words and measuring how good these embeddings are.

Quantitatively, we follow \newcite{luong13} in using the word similarity task,
specifically on the {\it Rare Word} dataset, to judge the learned representations for
complex words. The evaluation metric is the Spearman's correlation $\rho$
between similarity scores assigned by a model and by human annotators.
From the results in Table~\ref{t:word_sim}, we can see that source representations produced by
our hybrid\footnote{We look up the encoder embeddings for frequent words and build representations for
rare word from characters.}  models
are significantly better than those of the word-based one. It is noteworthy that our deep recurrent
character-level models can outperform the model of \cite{luong13}, which uses
recursive neural networks and requires a complex morphological analyzer, by a large
margin. Our performance is also competitive to the best Glove embeddings
\cite{pennington2014} which were trained on a much larger dataset.
\begin{table}[tbh!]
\centering
\resizebox{8cm}{!}{
\begin{tabular}{c|l|c|c|c}
\multicolumn{2}{c|}{{\bf System}} & Size & $|V|$ & \bf{$\rho$}\\ 
  \hline
\multicolumn{2}{l|}{\cite{luong13}} & 1B & 138K & 34.4 \\
  \hdashline
\multicolumn{2}{l|}{\multirow{2}{*}{Glove \cite{pennington2014}}} & 6B & 400K & 38.1 \\
\multicolumn{2}{l|}{} & 42B & 400K & \bf{47.8} \\
  \hline
\multicolumn{4}{c}{{\it Our NMT models}}\\
  \hline
\modelword{} & Word-based & 0.3B & 50K & 20.4 \\
  \hdashline
\modelsmall{} & Hybrid & 0.3B & 10K & 42.4 \\
  \hdashline
\model{} & Hybrid & 0.3B & 50K & \bi{47.1} \\
\end{tabular}
}
\caption{{\bf Word similarity task} -- shown are Spearman's correlation
$\rho$ on the {\it Rare Word} dataset
of various models (with different vocab sizes $|V|$). 
} 
\label{t:word_sim}
\end{table}

Qualitatively, we visualize embeddings produced by the hybrid model \model{} for
selected words in the Rare Word dataset.
Figure~\ref{f:visual} shows the two-dimensional representations of words
computed by the
Barnes-Hut-SNE algorithm \cite{bhsne}.\footnote{We run Barnes-Hut-SNE algorithm
over a set of 91 words, but filter out 27 words for displaying clarity.} It is extremely interesting to observe that
words are clustered together not only by the word structures but also by
the meanings. For example, in the top-left box,
the {\it character}-based representations for \word{loveless}, \word{spiritless}, \word{heartlessly}, and \word{heartlessness} are nearby,
but clearly separated into two groups. Similarly, in the center boxes, {\it
word}-based embeddings of
\word{acceptable}, \word{satisfactory}, \word{unacceptable}, and \word{unsatisfactory}, are
close by but separated by meanings. Lastly, the remaining boxes demonstrate that our
character-level models are able to build representations comparable to the
word-based ones, e.g., \word{impossibilities} vs.\ \word{impossible} and \word{antagonize}
vs.\ \word{antagonist}. All of this evidence strongly supports that the source
character-level models are useful and effective.

\begin{table*}
\centering
\resizebox{16cm}{!}{
\begin{tabular}{c|c|p{16cm}}
\multirow{7}{*}{1} & source & The author \source{Stephen Jay Gould} died 20 years after
\source{diagnosis} . \\
& human & Autor \correct{Stephen Jay Gould} zem\u{r}el 20 let po
\correct{diagn\'oze} . \\
  \cline{2-3}
& \multirow{2}{*}{\it{word}} & Autor Stephen Jay \unk{} zem\u{r}el 20 let po
\unk{} . \\
&  & Autor \correct{Stephen Jay Gould} zem\u{r}el 20 let po \wrong{po} .\\
  \cline{2-3}
& \it{char} & Autor \wrong{Stepher Stepher} zem\u{r}el 20 let po
\correct{diagn\'oze} . \\
  \cline{2-3}
& \multirow{2}{*}{\it{hybrid}} & Autor \unk{} \unk{} \unk{} zem\u{r}el 20 let po
\unk{}. \\
&  & Autor \correct{Stephen Jay Gould} zem\u{r}el 20 let po \correct{diagn\'oze} .\\
  \hline
  \hline
\multirow{7}{*}{2} & source & As the Reverend \source{Martin Luther King
Jr.} said \source{fifty years ago} :\\
& human & Jak \correct{p\u{r}ed pades\'ati lety} \u{r}ekl reverend \correct{Martin
Luther King Jr} . : \\
  \cline{2-3}
& \multirow{2}{*}{\it{word}} & Jak \u{r}ekl reverend Martin \unk{} King \unk{}
p\u{r}ed pades\'ati lety : \\
&  & Jak \u{r}ekl reverend \correct{Martin Luther King} \wrong{\u{r}ekl}
\close{p\u{r}ed pades\'ati lety} : \\
  \cline{2-3}
& \it{char} & Jako reverend \correct{Martin Luther} \wrong{kr\'al \u{r}\'ikal}
\close{p\u{r}ed pades\'ati lety} : \\
  \cline{2-3}
& \multirow{2}{*}{\it{hybrid}} & Jak p\u{r}ed \unk{} lety \u{r}ekl \unk{} Martin
\unk{} \unk{} \unk{} : \\
&  & Jak \correct{p\u{r}ed pades\'ati lety} \u{r}ekl reverend \correct{Martin
Luther King} \close{Jr.} : \\
  \hline
  \hline
\multirow{7}{*}{3} & source & Her \source{11-year-old} daughter , \source{Shani Bart} , said it felt a " little bit
\source{weird} " [..] back to school . \\ 
& human & Jej\'{i} \correct{jeden\'{a}ctilet\'{a}} dcera \correct{Shani Bartov\'{a}} prozradila
, \u{z}e " je to trochu \correct{zvl\'{a}\u{s}tn\'{i}} " [..] znova do
\u{s}koly . \\ 
  \cline{2-3}
& \multirow{2}{*}{\it{word}} & Jej\'i \unk{} dcera \unk{} \unk{} \u{r}ekla , \u{z}e je to " trochu
divn\'e " , [..] vrac\'i do \u{s}koly .\\ 
&  & Jej\'i \wrong{11-year-old} dcera \correct{Shani} \wrong{,} \u{r}ekla , \u{z}e je to " trochu
\close{divn\'e} " , [..] vrac\'i do \u{s}koly . \\ 
  \cline{2-3}
& \it{char} & Jej\'i \correct{jeden\'actilet\'a} dcera , \correct{Shani
Bartov\'a} , \u{r}\'ikala ,
\u{z}e c\'it\'i trochu \close{divn\u{e}} , [..] vr\'atila do \u{s}koly .\\ 
  \cline{2-3}
& \multirow{2}{*}{\it{hybrid}} & Jej\'i \unk{} dcera , \unk{} \unk{} , \u{r}ekla , \u{z}e c\'it\'i " trochu
\unk{} " , [..] vr\'atila do \u{s}koly .\\ 
&  & Jej\'i \correct{jeden\'actilet\'a} dcera , \wrong{Graham} \close{Bart} , \u{r}ekla , \u{z}e c\'it\'i " trochu
\close{divn\'y} " , [..] vr\'atila do \u{s}koly . \\ 
\end{tabular}
}
\caption{{\bf Sample translations on newstest2015} -- 
for each example, we show the {\it source}, {\it human} translation, and
translations of the following NMT systems: {\it word} model \modelword{},
{\it char} model \modelchar{}, and {\it hybrid} model \modelsmall{}. We show the
translations before replacing \unk{} tokens (if any) for the word-based 
and hybrid models. The following formats are used to highlight
\correct{correct}, \wrong{wrong}, and \close{close} translation segments.}
\label{t:sample}
\end{table*}

\subsection{Sample Translations}
\label{subsec:samples}

We show in Table~\ref{t:sample} sample translations between various systems. 
In the first example, our hybrid model translates perfectly. The word-based
model fails to translate \word{diagnosis} because the second \unk{} was incorrectly
aligned to the word \word{after}. The character-based model, on the other hand,
makes a mistake in translating names.

For the second example, the hybrid model surprises us when it can capture
the long-distance reordering of \word{fifty years ago} and \word{p\u{r}ed
pades\'ati lety} while the other two models do not. The word-based model
translates \word{Jr.} inaccurately due to the incorrect alignment between the
second \unk{} and the word \word{said}. The
character-based model literally translates the name \word{King} into \word{kr\'al}
which means \word{king}.

Lastly, both the character-based and hybrid models impress us by
their ability to translate compound words exactly, e.g., \word{11-year-old} and
\word{jeden\'actilet\'a}; whereas the identity copy
strategy of the word-based model fails.
Of course, our hybrid model does make mistakes, e.g., it fails to translate the name
\word{Shani Bart}. 
Overall, these examples highlight how challenging translating
into Czech is and that being able to translate at the character level helps
improve the quality.

\section{Conclusion}
\label{sec:conclude}
We have proposed a novel {\it hybrid} architecture that combines the strength
of both word- and character-based models. Word-level models are fast to train
and offer high-quality translation; whereas, character-level models help achieve
the goal of open vocabulary NMT. 
We have demonstrated these two aspects through our experimental results and
translation examples.

Our best hybrid model has surpassed the performance of both the best word-based
NMT system and the best non-neural model to establish a new state-of-the-art result for 
English-Czech translation in WMT'15 with $\ensbleu{}$ BLEU.
Moreover, we have succeeded in replacing the standard unk replacement technique
in NMT with our character-level components, yielding an improvement of 
+$\gain{}$ BLEU points. Our analysis has shown that our model has the ability to
not only generate well-formed words for
Czech, a highly inflected language with an enormous and complex vocabulary, but
also build accurate representations for English source words.

Additionally, we have demonstrated the potential of purely character-based
models in producing good translations;
they have outperformed past word-level NMT models. For future work, we hope to be able to improve the memory usage and
speed of purely character-based models.

\section*{Acknowledgments}
This work was partially supported by NSF Award IIS-1514268 and by a gift from Bloomberg L.P.
We thank Dan Jurafsky, Andrew Ng, and Quoc Le for earlier feedback on the work,
as well as Sam Bowman, Ziang Xie, and Jiwei Li for their valuable comments on
the paper draft.
Lastly, we thank NVIDIA Corporation for the donation of
Tesla K40 GPUs as well as Andrew Ng and his group for letting us use their computing
resources.

\bibliography{acl2016}

\begin{thebibliography}{}

\bibitem[\protect\citename{Bahdanau \bgroup et al.\egroup }2015]{bog15}
Dzmitry Bahdanau, Kyunghyun Cho, and Yoshua Bengio.
\newblock 2015.
\newblock Neural machine translation by jointly learning to align and
  translate.
\newblock In {\em ICLR}.

\bibitem[\protect\citename{Bahdanau \bgroup et al.\egroup }2016]{bahdanau16}
Dzmitry Bahdanau, Jan Chorowski, Dmitriy Serdyuk, Philemon Brakel, and Yoshua
  Bengio.
\newblock 2016.
\newblock End-to-end attention-based large vocabulary speech recognition.
\newblock In {\em ICASSP}.

\bibitem[\protect\citename{Ballesteros \bgroup et al.\egroup
  }2015]{ballesteros15}
Miguel Ballesteros, Chris Dyer, and Noah~A. Smith.
\newblock 2015.
\newblock Improved transition-based parsing by modeling characters instead of
  words with {LSTMs}.
\newblock In {\em EMNLP}.

\bibitem[\protect\citename{Bojar and Tamchyna}2015]{bojar15wmt}
Ond\u{r}ej Bojar and Ale\u{s} Tamchyna.
\newblock 2015.
\newblock {CUNI in WMT15: Chimera Strikes Again}.
\newblock In {\em WMT}.

\bibitem[\protect\citename{Chan \bgroup et al.\egroup }2016]{chan16}
William Chan, Navdeep Jaitly, Quoc~V. Le, and Oriol Vinyals.
\newblock 2016.
\newblock Listen, attend and spell.
\newblock In {\em ICASSP}.

\bibitem[\protect\citename{Cho \bgroup et al.\egroup }2014]{cho14}
Kyunghyun Cho, Bart van Merrienboer, Caglar Gulcehre, Fethi Bougares, Holger
  Schwenk, and Yoshua Bengio.
\newblock 2014.
\newblock Learning phrase representations using {RNN} encoder-decoder for
  statistical machine translation.
\newblock In {\em EMNLP}.

\bibitem[\protect\citename{dos Santos and Zadrozny}2014]{santos14}
C{\'{\i}}cero~Nogueira dos Santos and Bianca Zadrozny.
\newblock 2014.
\newblock Learning character-level representations for part-of-speech tagging.
\newblock In {\em ICML}.

\bibitem[\protect\citename{Hochreiter and Schmidhuber}1997]{lstm97}
Sepp Hochreiter and J\"{u}rgen Schmidhuber.
\newblock 1997.
\newblock Long short-term memory.
\newblock 9(8):1735--1780.

\bibitem[\protect\citename{Jean \bgroup et al.\egroup }2015a]{jean15}
S\'{e}bastien Jean, Kyunghyun Cho, Roland Memisevic, and Yoshua Bengio.
\newblock 2015a.
\newblock On using very large target vocabulary for neural machine translation.
\newblock In {\em ACL}.

\bibitem[\protect\citename{Jean \bgroup et al.\egroup }2015b]{jean15wmt}
S\'{e}bastien Jean, Orhan Firat, Kyunghyun Cho, Roland Memisevic, and Yoshua
  Bengio.
\newblock 2015b.
\newblock Montreal neural machine translation systems for {WMT}'15.
\newblock In {\em WMT}.

\bibitem[\protect\citename{Jozefowicz \bgroup et al.\egroup }2016]{rafal16}
Rafal Jozefowicz, Oriol Vinyals, Mike Schuster, Noam Shazeer, and Yonghui Wu.
\newblock 2016.
\newblock Exploring the limits of language modeling.

\bibitem[\protect\citename{Kalchbrenner and Blunsom}2013]{kal13}
Nal Kalchbrenner and Phil Blunsom.
\newblock 2013.
\newblock Recurrent continuous translation models.
\newblock In {\em EMNLP}.

\bibitem[\protect\citename{Kim \bgroup et al.\egroup }2016]{kim16}
Yoon Kim, Yacine Jernite, David Sontag, and Alexander~M. Rush.
\newblock 2016.
\newblock Character-aware neural language models.
\newblock In {\em AAAI}.

\bibitem[\protect\citename{Li \bgroup et al.\egroup }2015]{li15}
Jiwei Li, Minh-Thang Luong, and Dan Jurafsky.
\newblock 2015.
\newblock A hierarchical neural autoencoder for paragraphs and documents.
\newblock In {\em ACL}.

\bibitem[\protect\citename{Liang \bgroup et al.\egroup }2006]{liang06alignment}
Percy Liang, Ben Taskar, and Dan Klein.
\newblock 2006.
\newblock Alignment by agreement.
\newblock In {\em NAACL}.

\bibitem[\protect\citename{Ling \bgroup et al.\egroup }2015a]{ling15function}
Wang Ling, Chris Dyer, Alan~W. Black, Isabel Trancoso, Ramon Fermandez, Silvio
  Amir, Lu\'{i}s Marujo, and Tiago Lu\'{i}s.
\newblock 2015a.
\newblock Finding function in form: Compositional character models for open
  vocabulary word representation.
\newblock In {\em EMNLP}.

\bibitem[\protect\citename{Ling \bgroup et al.\egroup }2015b]{ling15char}
Wang Ling, Isabel Trancoso, Chris Dyer, and Alan Black.
\newblock 2015b.
\newblock Character-based neural machine translation.

\bibitem[\protect\citename{Luong and Manning}2015]{luong15iwslt}
Minh-Thang Luong and Christopher~D. Manning.
\newblock 2015.
\newblock Stanford neural machine translation systems for spoken language
  domain.
\newblock In {\em IWSLT}.

\bibitem[\protect\citename{Luong \bgroup et al.\egroup }2013]{luong13}
Minh-Thang Luong, Richard Socher, and Christopher~D. Manning.
\newblock 2013.
\newblock Better word representations with recursive neural networks for
  morphology.
\newblock In {\em CoNLL}.

\bibitem[\protect\citename{Luong \bgroup et al.\egroup }2015a]{luong15attn}
Minh-Thang Luong, Hieu Pham, and Christopher~D. Manning.
\newblock 2015a.
\newblock Effective approaches to attention-based neural machine translation.
\newblock In {\em EMNLP}.

\bibitem[\protect\citename{Luong \bgroup et al.\egroup }2015b]{luong15}
Minh-Thang Luong, Ilya Sutskever, Quoc~V. Le, Oriol Vinyals, and Wojciech
  Zaremba.
\newblock 2015b.
\newblock Addressing the rare word problem in neural machine translation.
\newblock In {\em ACL}.

\bibitem[\protect\citename{Papineni \bgroup et al.\egroup
  }2002]{Papineni02bleu}
Kishore Papineni, Salim Roukos, Todd Ward, and Wei jing Zhu.
\newblock 2002.
\newblock Bleu: a method for automatic evaluation of machine translation.
\newblock In {\em ACL}.

\bibitem[\protect\citename{Pennington \bgroup et al.\egroup
  }2014]{pennington2014}
Jeffrey Pennington, Richard Socher, and Christopher~D. Manning.
\newblock 2014.
\newblock Glove: Global vectors for word representation.
\newblock In {\em EMNLP}.

\bibitem[\protect\citename{Pham \bgroup et al.\egroup }2014]{pham2014dropout}
Vu~Pham, Th\'{e}odore Bluche, Christopher Kermorvant, and J\'{e}r\^{o}me
  Louradour.
\newblock 2014.
\newblock Dropout improves recurrent neural networks for handwriting
  recognition.
\newblock In {\em ICFHR}.

\bibitem[\protect\citename{Popovi\'{c}}2015]{chrf}
Maja Popovi\'{c}.
\newblock 2015.
\newblock {chrF: character n-gram F-score for automatic MT evaluation}.
\newblock In {\em WMT}.

\bibitem[\protect\citename{Sennrich \bgroup et al.\egroup }2016]{sennrich16sub}
Rico Sennrich, Barry Haddow, and Alexandra Birch.
\newblock 2016.
\newblock Neural machine translation of rare words with subword units.
\newblock In {\em ACL}.

\bibitem[\protect\citename{Sutskever \bgroup et al.\egroup }2014]{sutskever14}
Ilya Sutskever, Oriol Vinyals, and Quoc~V. Le.
\newblock 2014.
\newblock Sequence to sequence learning with neural networks.
\newblock In {\em NIPS}.

\bibitem[\protect\citename{van~der Maaten}2013]{bhsne}
Laurens van~der Maaten.
\newblock 2013.
\newblock {Barnes-Hut-SNE}.
\newblock In {\em ICLR}.

\bibitem[\protect\citename{Zaremba \bgroup et al.\egroup }2014]{zaremba14}
Wojciech Zaremba, Ilya Sutskever, and Oriol Vinyals.
\newblock 2014.
\newblock Recurrent neural network regularization.
\newblock abs/1409.2329.

\bibitem[\protect\citename{Zhang \bgroup et al.\egroup }2015]{zhang15}
Xiang Zhang, Junbo Zhao, and Yann LeCun.
\newblock 2015.
\newblock Character-level convolutional networks for text classification.
\newblock In {\em NIPS}.

\end{thebibliography}
\bibliographystyle{acl2016}

\end{document}